\documentclass[letterpaper]{article} 
\usepackage{aaai25}  
\usepackage{times}  
\usepackage{helvet}  
\usepackage{courier}  
\usepackage[hyphens]{url}  
\usepackage{graphicx} 
\usepackage{subcaption}
\usepackage{natbib}  
\usepackage{caption} 
\usepackage{algorithm}
\usepackage{algorithmic}
\usepackage{newfloat}
\usepackage{listings}
\DeclareCaptionStyle{ruled}{labelfont=normalfont,labelsep=colon,strut=off} 
\floatstyle{ruled}
\newfloat{listing}{tb}{lst}{}
\floatname{listing}{Listing}
\title{AgentDNS: A Root Domain Naming System for LLM Agents}
\author{
    Enfang Cui\textsuperscript{\rm 1},
    Yujun Cheng\textsuperscript{\rm 2},
    Rui She\textsuperscript{\rm 1},
    Dan Liu\textsuperscript{\rm 1},
    Zhiyuan Liang\textsuperscript{\rm 1},
    Minxin Guo\textsuperscript{\rm 1},
    Tianzheng Li\textsuperscript{\rm 1},
    Qian Wei\textsuperscript{\rm 1},
    Wenjuan Xing\textsuperscript{\rm 1},
    Zhijie Zhong\textsuperscript{\rm 3,4}
}
\affiliations{
    \textsuperscript{\rm 1}China Telecom Research Institute, Beijing, China\\
    \textsuperscript{\rm 2}School of Intelligence Science and Technology, University of Science and Technology Beijing, Beijing, China\\
    \textsuperscript{\rm 3}School of Future Technology, South China University of Technology, Guangzhou, Guangdong, China\\
    \textsuperscript{\rm 4}Pengcheng Laboratory, Shenzhen, Guangdong, China\\
    
    cuief@chinatelecom.cn, yjcheng@tsinghua.edu.cn, 
    \{sher, liud11, liangzy17, guomx3, litz4, weiq12, xingwj\}@chinatelecom.cn,
    csemor@mail.scut.edu.cn
}

\usepackage{bibentry}

\begin{document}

\maketitle

\begin{abstract}
The rapid evolution of Large Language Model (LLM) agents has highlighted critical challenges in cross-vendor service discovery, interoperability, and communication. Existing protocols like model context protocol and agent-to-agent protocol have made significant strides in standardizing interoperability between agents and tools, as well as communication among multi-agents. However, there remains a lack of standardized protocols and solutions for service discovery across different agent and tool vendors. In this paper, we propose AgentDNS, a root domain naming and service discovery system designed to enable LLM agents to autonomously discover, resolve, and securely invoke third-party agent and tool services across organizational and technological boundaries. Inspired by the principles of the traditional DNS, AgentDNS introduces a structured mechanism for service registration, semantic service discovery, secure invocation, and unified billing. We detail the architecture, core functionalities, and use cases of AgentDNS, demonstrating its potential to streamline multi-agent collaboration in real-world scenarios. The source code will be published on {\textbf{https://github.com/agentdns}}. 
\end{abstract}

%

\section{Introduction}
In recent years, LLM agent \cite{luo2025large} technology has been reshaping industries at an unprecedented pace. Leveraging natural language understanding, multi-modal interaction capabilities, and complex task orchestration, LLM agents have penetrated core sectors such as education \cite{chu2025llm}, finance \cite{ding2024large}, and academic \cite{alzubi2025open,chen2024mindsearch}, etc., driving intelligent transformation across domain-specific workflows. Market research indicates that the global LLM agent market is projected to exceed \$50 billion by 2030 \cite{marketsandmarkets_ai_agents}. This growth stems from the flexibility of ``natural language as instructions''---users can describe requirements in everyday language, enabling agents to automatically invoke toolchains, parse heterogeneous data, and complete end-to-end tasks. The dual drive of technological advancements and commercial adoption has positioned LLM agents as a foundational infrastructure for enterprise digital transformation.

\begin{figure}[t]
\centering
\includegraphics[width=0.9\linewidth]{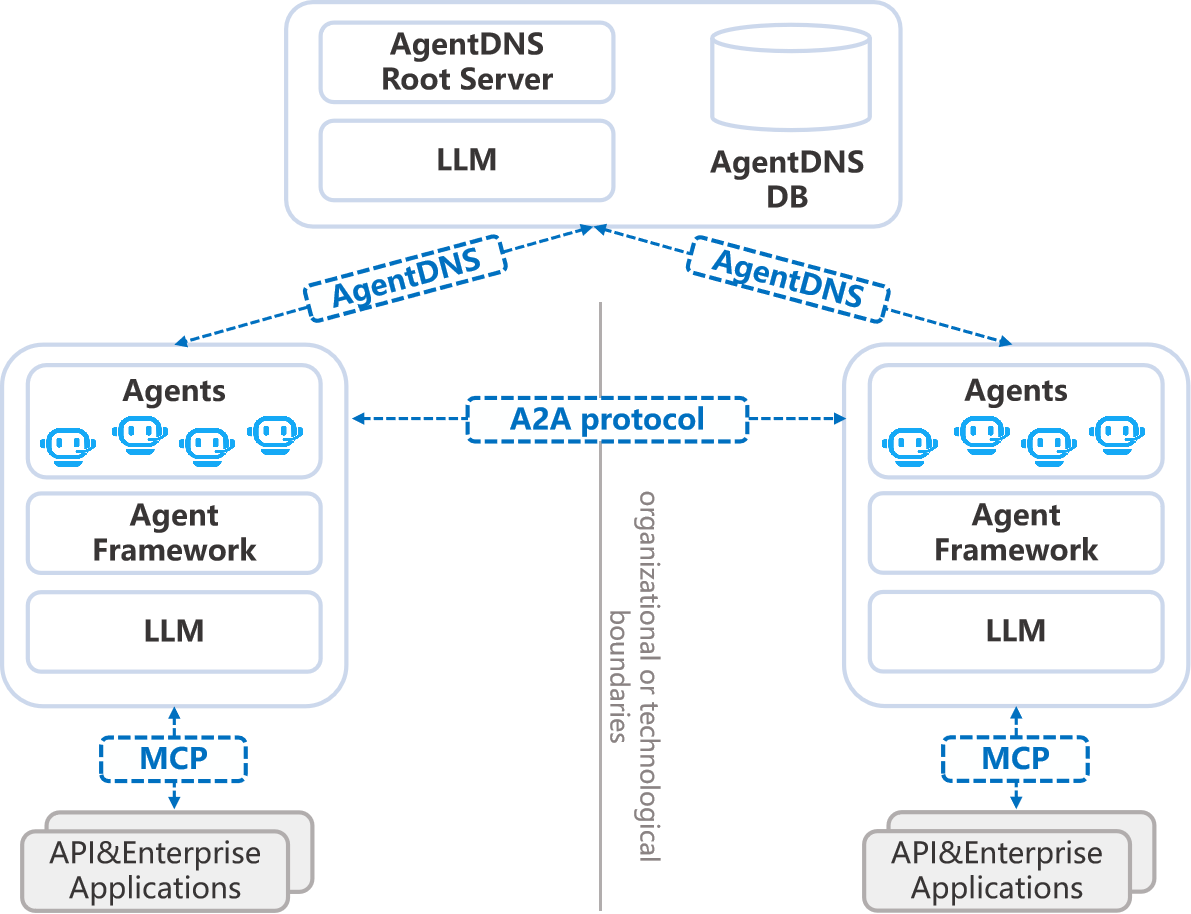}
\caption{AgentDNS system and its relationship with A2A and MCP Protocols.}
\label{fig:arch}
\end{figure}

The current focus of technological evolution is shifting from enhancing single-agent capabilities to building multi-agent collaborative systems \cite{guo2024large, han2024llm}, transcending the limitations of isolated intelligence. In industry practice, multiple agent interoperability and communication protocols \cite{yang2025survey} have been proposed, such as Anthropic's model context protocol (MCP) \cite{hou2025model}, which standardizes tool invocation interfaces of diverse toolchains.  Meanwhile, Google's Agent-to-Agent (A2A) protocol \cite{googlea2a} establishes task delegation mechanisms between agents, supporting cross-organizational workflow orchestration. The maturation of technical standards and open-source frameworks is accelerating the development of an open, scalable multi-agent collaboration ecosystem.

In the future, we envision a world where agents can autonomously discover, communicate, and collaborate with one another without human intervention. Although protocols like MCP and A2A have effectively facilitated communication and collaboration between agents and external tools, as well as between agents themselves, there is still a lack of standardized protocols and systems for cross-vendor service naming, service discovery, authentication, and billing. As a result, agent-to-agent collaboration still demands significant manual effort, preventing the realization of true autonomous cooperation. The specific challenges are as follows:

\begin{itemize}
    \item \textbf{The Service Discovery Challenge:}
    LLM agents typically generate an action plan, where each action may require calling external tools or agent services. However, currently, services from different vendors are not standardized in naming or management, which forces developers to manually maintain service information for each tool or agent. This lack of standardization makes it impossible for LLM agents to autonomously discover other agents or tool services, hindering seamless integration and collaboration between agents across different platforms.
    \item \textbf{The Interoperability Challenge:}
    Currently, different vendors’ agents or tool services support various interoperability or communication protocols, with typical examples including MCP, A2A, and ANP\cite{anp2025}. We anticipate that more interoperability protocols will emerge in the future. However, LLM agents are unable to autonomously recognize and adapt to these differences, meaning they still require manual configuration and management. This lack of flexibility in handling diverse protocols limits seamless agent-to-agent and agent-to-tool communication across platforms.
    \item \textbf{The Authentication and Billing Challenge:}
    Cross-vendor collaboration is further complicated by security and authentication challenges. Each service provider typically requires proprietary API keys, necessitating manual configuration of multiple authentication credentials for agents. This adds significant overhead and disrupts seamless integration. In addition, billing systems are fragmented across vendors, requiring manual intervention for payment setup. As a result, agents are unable to autonomously discover and invoke new third-party paid agents or tool services without manual configuration.
\end{itemize}

To address these challenges, we propose \textbf{AgentDNS}, the first root domain naming and resolution system designed for LLM agents. Inspired by the Internet’s Domain Name System (DNS), AgentDNS introduces a unified namespace (\textbf{agentdns://}), natural language-based service discovery, and unified authentication and billing. As shown in Fig.\ref{fig:arch}, AgentDNS is compatible with protocols such as MCP and A2A, allowing them to coexist seamlessly. With AgentDNS, agents can autonomously discover services, authenticate, and interoperate seamlessly across organizational boundaries. AgentDNS redefines the multi-agent collaboration ecosystem through four key functions:

\begin{itemize}
    \item \textbf{Unified Namespace with Semantic Information:} AgentDNS introduces a semantically rich naming scheme (e.g., agentdns://org/category/name) for agents and tool services, decoupling service identifier name from physical addresses such as URLs. This also enables the semantic embedding of agent capabilities into their identifier name, facilitating more efficient classification and retrieval of agent and tool services.
    \item \textbf{Natural Language-Driven Service Discovery}: Agents can interact with the AgentDNS root service using natural language queries to discover third-party agents or tool services. They can obtain the corresponding service identifier names and related metadata, including physical addresses, capabilities, and communication protocol, etc. Agents can also dynamically request the AgentDNS root service to resolve an identifier name and retrieve the latest metadata as needed.
    \item \textbf{Protocol-Aware Interoperability}: AgentDNS enables agents to dynamically discover the supported interoperability or communication protocols of third-party agents and tool services by resolving their identifier names into metadata. This metadata includes not only network addresses and capabilities, but also the specific protocols (e.g., MCP, A2A, ANP) each service supports. Based on this information, agents can autonomously select and adapt to the appropriate protocol for communication, eliminating the need for manual configuration.
    \item \textbf{Unified Authentication and Billing}: AgentDNS replaces fragmented API keys with a single-sign-on mechanism. Agents authenticate once with the AgentDNS root server to obtain time-bound access tokens, valid across all registered services. For billing, AgentDNS serves as a unified billing platform: users pre-fund accounts, usage costs are tracked and deducted in real-time, and payments are automatically settled across vendors. This enables transparent billing and autonomous access to paid services by agents.
\end{itemize}

\section{Related Work}
\subsection{LLM Agents}
LLM agents have rapidly emerged as a pivotal research frontier in artificial intelligence, driven by their transformative potential to bridge human-AI collaboration and autonomous problem-solving. In the industrial, several LLM agents have been launched, such as Deep Research \cite{openai2025deepresearch}, Manus \cite{manus2025}, and Cursor \cite{cursor2025}, etc. Unlike traditional AI systems constrained by predefined rules, LLM agents leverage the general-purpose reasoning, contextual understanding, and multi-task capabilities of LLMs to dynamically adapt to complex environments. LLM agents have demonstrated broad application prospects across various fields. The future of LLM agents is expected to trend towards multi-agent collaboration. Researchers are increasingly interested in how to design efficient communication protocols and coordination mechanisms \cite{hou2025model, googlea2a, li2024improving, marro2024scalable} that enable seamless cooperation among agents. This collaborative approach is seen as a key direction for advancing the capabilities and applications of LLM agents in the coming years.
\subsection{Agent Interaction Protocols}
\subsubsection{Model Context Protocol}
The Model Context Protocol (MCP) \cite{hou2025model} is an open standard developed by Anthropic, designed to facilitate seamless interactions between LLM models and external tools, data sources, and services. Inspired by the concept of a universal adapter, MCP aims to simplify the integration process, much like how a USB-C port allows various devices to connect effortlessly. MCP operates on a client-server architecture. The AI application (such as a chatbot or an integrated development environment) acts as the host and runs an MCP client, while each external integration runs as an MCP server. The server exposes capabilities such as functions, data resources, or predefined prompts, and the client connects to it to utilize these capabilities. This design allows AI models to interact with external systems without directly accessing APIs, thereby enhancing security and reducing the complexity of custom integrations.

\begin{figure}[t]
\centering
\includegraphics[width=1\linewidth]{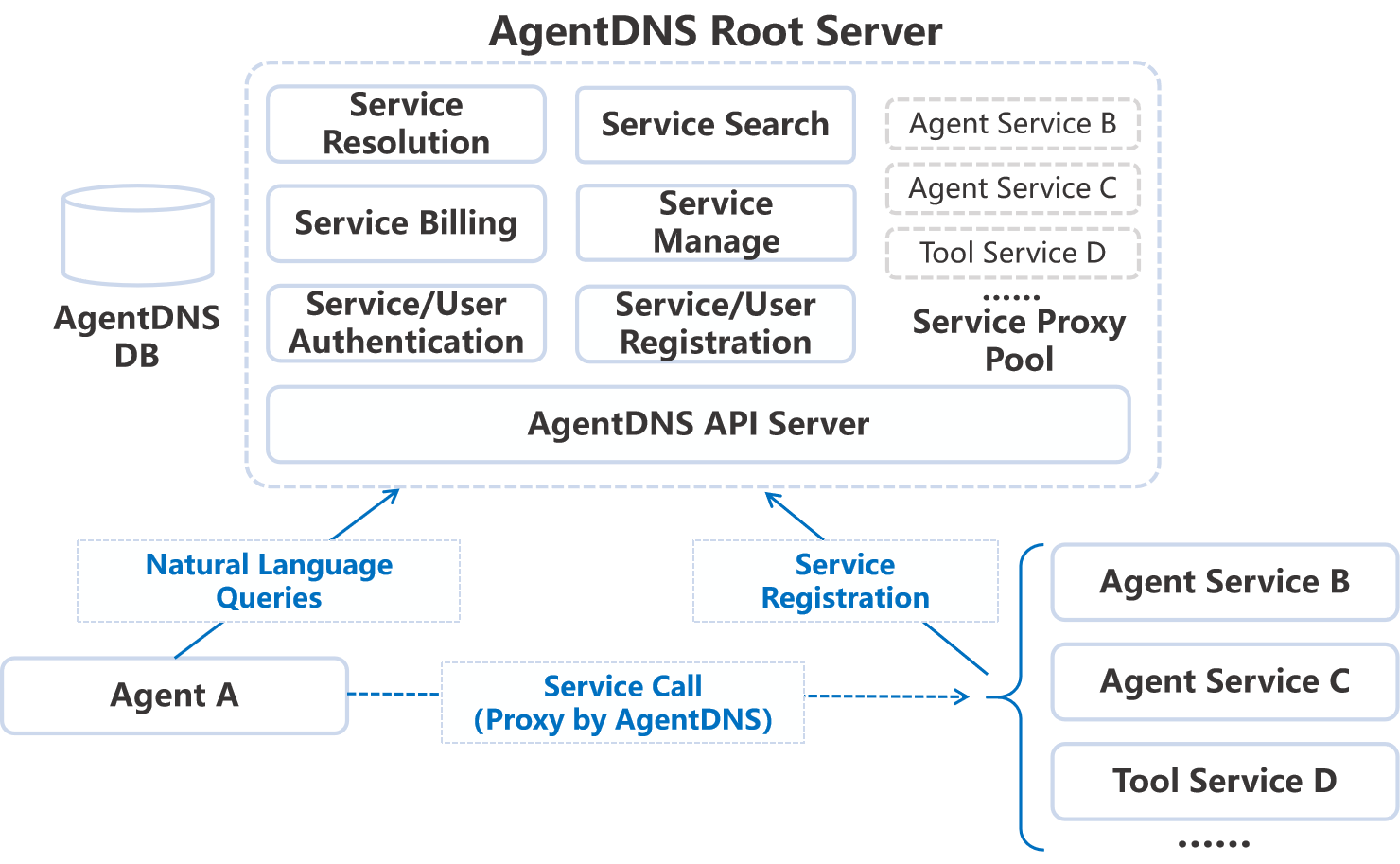}
\caption{AgentDNS system architecture.}
\label{fig:systemoverview}
\end{figure}

\subsubsection{Agent-to-Agent Protocol}
The Agent-to-Agent (A2A) protocol \cite{googlea2a} is introduced by Google, aimed at enabling seamless communication and collaboration between LLM agents, regardless of their underlying frameworks or vendors. A2A was developed in collaboration with over 50 technology partners, including major companies like Atlassian, Salesforce, SAP, and MongoDB. This protocol uses HTTP-based APIs and JSON data format, ensuring compatibility and ease of integration with existing enterprise IT systems. It supports various communication patterns, including request-response, event-based communication, and streaming data exchange. A2A complements protocols like MCP, which focuses on providing tools and context for agents. A2A focuses on agent-to-agent communication, allowing them to work together more effectively.

\subsection{Domain Naming System}
The Domain Name System (DNS) \cite{danzig1992analysis, cheshire2013rfc6763} serves as the critical naming and discovery infrastructure for the human internet, translating memorable domain names (e.g., example.com) into physical addresses (IP addresses) through its hierarchical, decentralized architecture. While DNS effectively decouples human-readable names from machine-level addressing, its design proves inadequate for the emerging agent Internet where LLM agents require autonomous service discovery and interoperability. Traditional DNS lacks three critical capabilities essential for agent ecosystems: service discovery through natural language, querying service metadata beyond physical addresses (including capabilities, protocols, etc.), and unified authentication and billing. These limitations necessitate AgentDNS-a purpose-built system that preserves DNS's core benefits of naming and resolution while introducing agent-specific innovations. 

\section{AgentDNS Method}
\subsection{AgentDNS System Overview}
AgentDNS is a root service system for agent service naming, discovery, and resolution, enabling seamless service discovery, cross-vendor interoperability, unified authentication and billing. As shown in Fig. \ref{fig:systemoverview}, the AgentDNS root server is the central hub of the entire system, which connects agent users (e.g., Agent A) with service providers (e.g., Agent Service B, Tool Service D). The AgentDNS root server comprises components including service registration, service proxy pool, service management, service search, service resolution, billing, authentication, etc. The core components are as follows:
\begin{itemize}
\item Service Registration Component: Agent or tool service vendors register their services through this component. The process begins with organization registration, where developers first create an organization account. Under the organization’s domain, they then setup a service category and name to generate a globally unique service identifier name (e.g., agentdns://org/category/name). Concurrently, developers must submit service metadata to bind with the identifier, including network addresses (e.g., endpoints, URLs), supported interoperability protocols (e.g., MCP, A2A), detailed service capabilities, etc. This metadata is securely stored in the AgentDNS database. During subsequent service discovery or resolution phases, the system performs semantic matching by analyzing the identifier’s category and the metadata. This ensures precise retrieval of services aligned with an agent’s operational requirements.

\item Service Proxy Pool: After a vendor registers a service, AgentDNS creates a corresponding service proxy within the Service Proxy Pool. This proxy acts as an intermediary, forwarding service invocation requests from user agents to the vendor’s actual service endpoint. In other words, the user agent sends a service request to the AgentDNS root server, which then routes the request to the appropriate vendor for execution. This design allows user agents to authenticate only once with AgentDNS, eliminating the need for separate registration and authentication with each individual vendor.

\item Service Search Component: User agents can send natural language queries to the AgentDNS root server to discover relevant third-party agents or tool services. This component interprets the query and performs intelligent retrieval using a combination of keyword matching and retrieval-augmented generation (RAG) \cite{gao2023retrieval} techniques. Based on the search results, it returns a list of top-k candidate services that match the agent’s intent. Each result includes the service identifier name, physical endpoint, supported communication protocols, capability descriptions, and pricing information. The user agent can then evaluate these candidates and choose the most appropriate one for execution. Once selected, the agent can directly invoke the service by using the appropriate protocol and access the physical endpoint with an authentication token issued by AgentDNS.

\begin{figure}[t]
\centering
\includegraphics[width=0.8\linewidth]{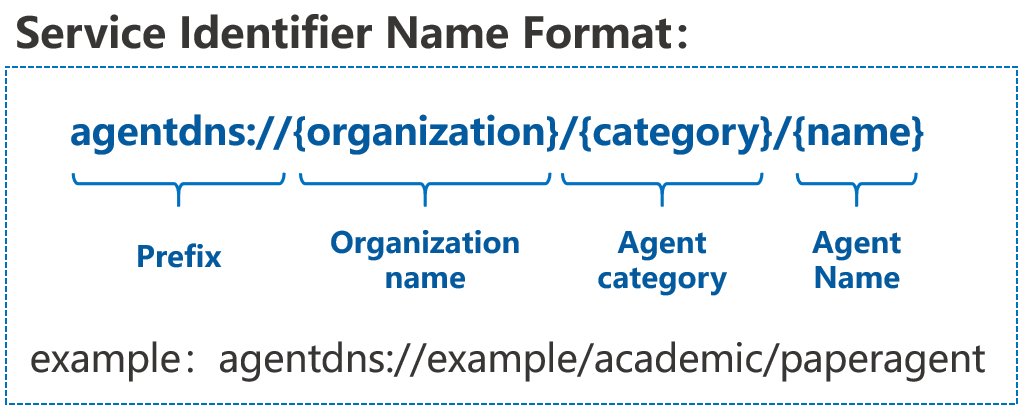}
\caption{AgentDNS service naming.}
\label{fig:serviceid}
\end{figure}

\item Service Resolution Component: User agents can cache service identifier names and, during subsequent invocations, dynamically request the AgentDNS root server to resolve these identifiers and get the latest metadata as needed.

\item Service Management Component: This component handles the lifecycle of these service proxies, including their creation, updates, and deletion, ensuring that the proxy infrastructure remains up-to-date and synchronized with the underlying services.

\item Service Billing Component: This component is responsible for tracking and processing service invocation costs. Users only need to settle payments directly with AgentDNS, which then handles the backend settlement with individual vendors. This design significantly simplifies the billing process for users by eliminating the need for managing multiple vendor-specific payment systems, enabling a streamlined and unified billing experience.

\item Authentication Component: This component handles identity verification and access control for both user agents and service providers. Instead of requiring agents to authenticate separately with each vendor, AgentDNS offers a unified authentication mechanism. User agents authenticate once with the AgentDNS root server and receive a time-bound access token. This token can be used to securely access any registered third-party service without additional logins. By centralizing authentication, this component ensures secure, efficient, and scalable access across a heterogeneous agent ecosystem, while also reducing the operational burden on both users and service vendors.

\end{itemize}

Together, these components form the backbone of AgentDNS, providing a unified framework that supports natural language-driven discovery, protocol-aware interoperability, trustless authentication, and unified billing—paving the way for truly autonomous multi-agent ecosystems. Next, we will provide a detailed introduction to AgentDNS’s service naming, service discovery, service resolution, and unified authentication and billing mechanisms.

\begin{figure}[t]
\centering
\begin{subfigure}{1\linewidth}
    \centering
    \includegraphics[width=0.9\linewidth]{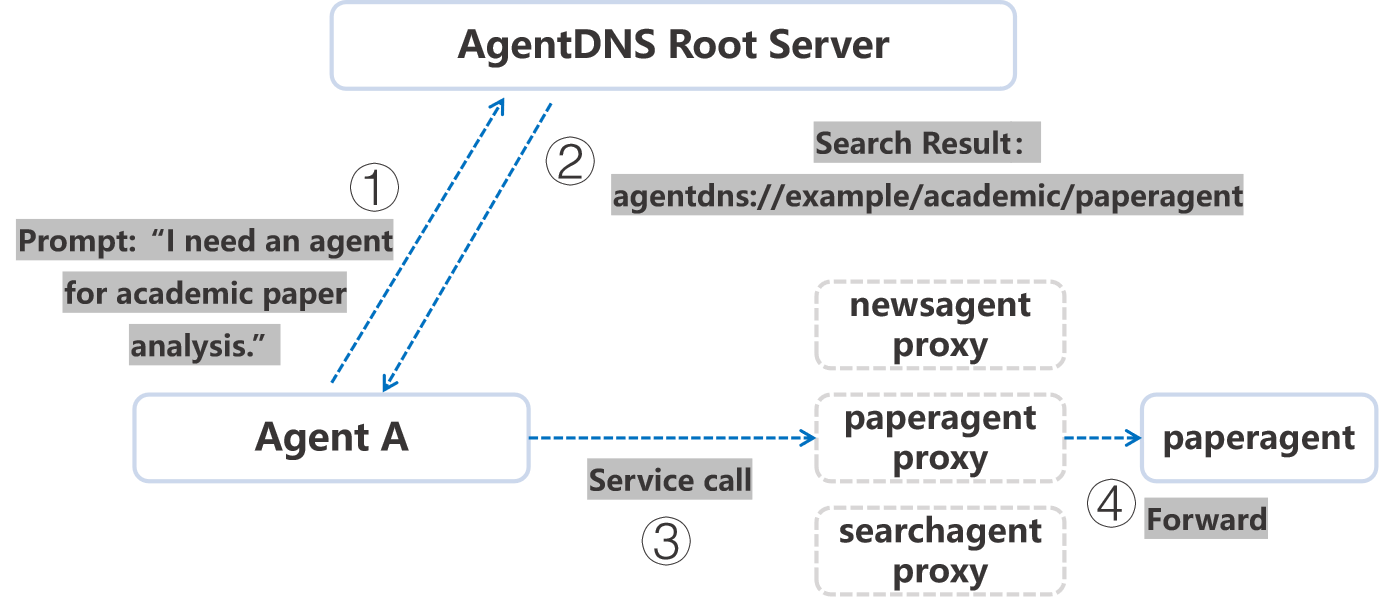}
    \caption{AgentDNS service discovery.}
    \label{fig:discover}
\end{subfigure}

\vspace{1em}  

\begin{subfigure}{1\linewidth}
    \centering
    \includegraphics[width=1\linewidth]{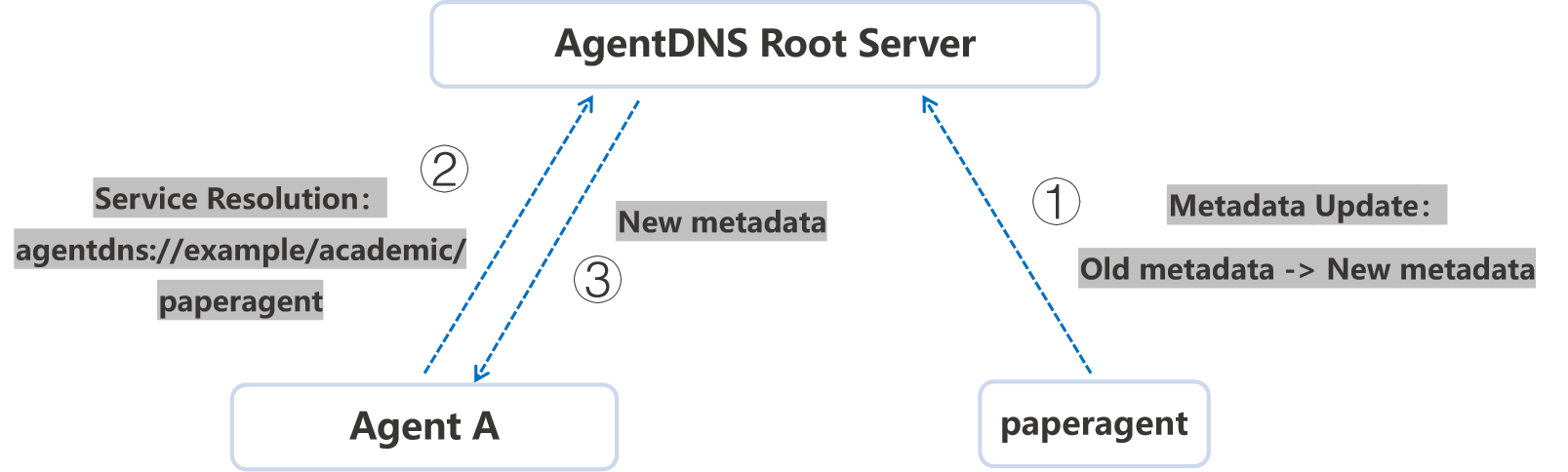}
    \caption{AgentDNS service resolution.}
    \label{fig:resolution}
\end{subfigure}

\caption{AgentDNS service discovery and resolution.}
\label{fig:discoverresolution}
\end{figure}

\begin{figure}[t]
\centering
\includegraphics[width=1\linewidth]{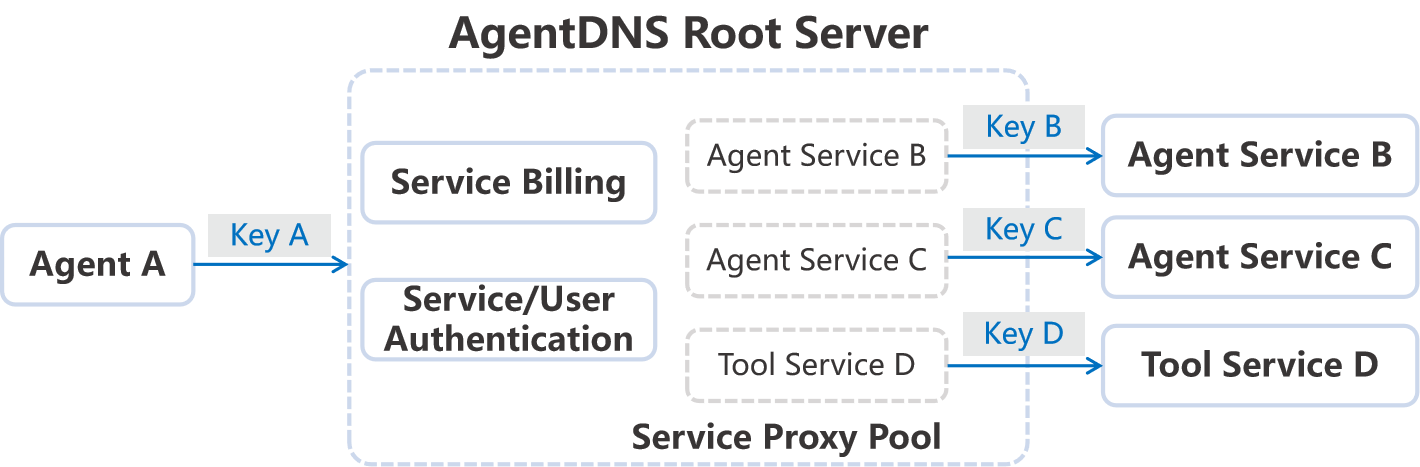}
\caption{AgentDNS unified authentication and billing.}
\label{fig:unify}
\end{figure}

\subsection{Service Naming}
The AgentDNS service naming system provides a structured and globally unique service identifier name for each registered agent or tool service. The identifier name follows the format as shown in Fig. \ref{fig:serviceid}. The organization represents the name of the registering entity, such as a company, university, or research lab. Each organization must go through a registration and verification process to ensure uniqueness and authenticity. The category denotes the functional domain or classification of the agent service. This can be chosen manually by the developer or automatically generated by AgentDNS, and it supports hierarchical structures—allowing for multi-level categories using slashes (e.g., academic/nlp/summarization). Finally, the name is the unique identifier for the specific agent within the organization and category. This name must be explicitly defined by the developer. Together, this structured naming convention ensures precise identification, facilitates organized discovery, and supports scalable service management within the AgentDNS ecosystem.

\begin{figure*}[t]
\centering
\includegraphics[width=1\linewidth]{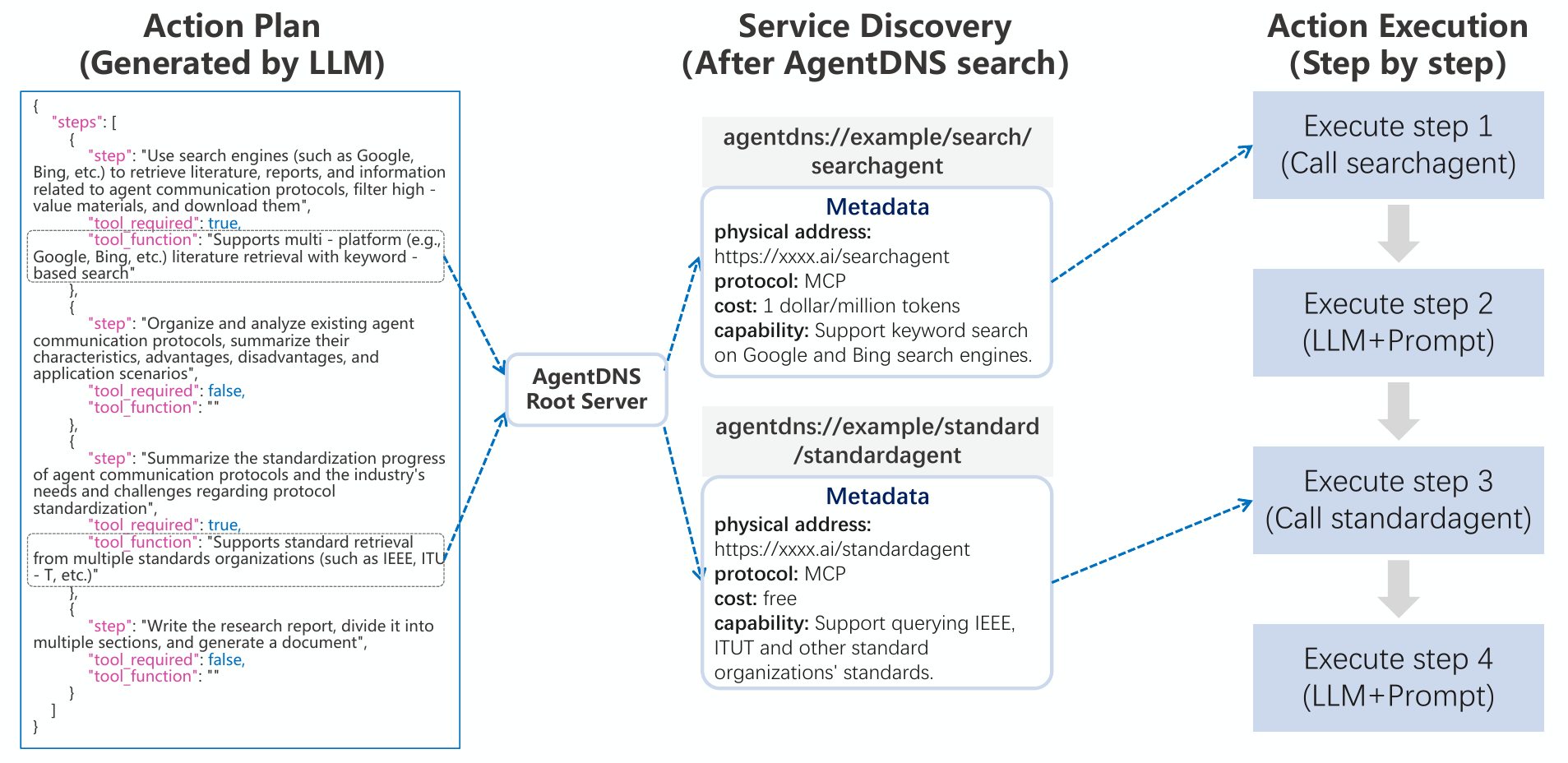}
\caption{AgentDNS case study.}
\label{fig:casestudy}
\end{figure*}

\subsection{Service Discovery}
The service discovery process is illustrated in Fig. \ref{fig:discover}. In step 1, Agent A initiates a natural language query to the AgentDNS root server, describing the desired service. In the example, Agent A is looking for an intelligent Agent capable of analyzing academic papers. In step 2, upon receiving the request, AgentDNS searches through its registry of available services to identify those with the required capabilities. It returns a list of service identifiers along with corresponding metadata, such as the proxy’s physical address, supported protocols, pricing information, and more. This discovery process employs a hybrid retrieval mechanism that combines keyword matching and RAG. Specifically, we construct a knowledge base using the capability descriptions of registered services. During service discovery, hybrid retrieval is performed over these capability descriptions to identify candidates that best match the user agent’s intent. In step 3, after receiving the service information, Agent A uses the appropriate protocol and an authentication token issued by AgentDNS to directly access the physical proxy address and initiate a service call. Finally, in step 4, the AgentDNS proxy forwards the request to the actual service endpoint hosted by the vendor, ensuring seamless interaction between Agent A and the service provider.

\subsection{Service Resolution}
As previously mentioned, user agents can cache service identifier names and request the AgentDNS root server for updated metadata when needed. This functionality helps reduce the frequency of accessing AgentDNS, improving response times and lowering operational costs. The service resolution process is illustrated in Fig. \ref{fig:resolution}. In step 1, agent vendors update the metadata associated with their agent services. In step 2, Agent A sends a resolution request to the AgentDNS root server, providing the cached service identifier name to retrieve the latest information. In step 3, AgentDNS locates the most recent metadata based on the identifier and returns it to Agent A, ensuring that the service invocation uses up-to-date information.

\subsection{Unified Authentication and Billing}
AgentDNS introduces a unified authentication and billing mechanism by acting as a proxy layer between user agents and third-party services. As shown in Fig. \ref{fig:unify}, when a user agent (e.g., Agent A) authenticates once with the AgentDNS root server using its own access key (Key A), it gains the ability to seamlessly invoke multiple external agent or tool services without needing to manage individual credentials for each provider. Internally, the AgentDNS root server maintains a service proxy pool that forwards user requests to the corresponding third-party services. For each third-party service, the proxy uses the appropriate authentication key (e.g., Key B, C, or D), which corresponds to the access control requirements of the service provider. This abstraction decouples the user agent from vendor-specific authentication logic. Moreover, billing is centralized: user agents are charged by AgentDNS based on their usage, while AgentDNS handles settlements with the respective third-party services. This model simplifies cross-vendor interoperability, enforces secure access, and enables consistent billing across a heterogeneous service ecosystem.

\section{AgentDNS Case Study}
In this section, we present a case study illustrating the interaction between an agent and the AgentDNS root server. The case demonstrates the complete agent workflow—from generating an action plan \cite{huang2024understanding}, to querying the AgentDNS root server for service discovery, and finally to executing the planned actions.

The full process is illustrated in Fig. \ref{fig:casestudy} After receiving a user request—such as “Help me research agent communication protocols and write a survey report”—the agent first invokes a LLM to generate an action plan. As shown in Fig. \ref{fig:casestudy}, the generated plan in this case is structured in JSON format and consists of multiple steps. Each step includes a description of its purpose, whether it requires a service, and a natural language description of the desired service functionality. These services correspond to third-party agent or tool services. For example, Step 1 requires a search service to retrieve relevant keywords, while Step 3 calls for a standards retrieval service to query documents from organizations like IEEE \cite{ieee2025} or ITU-T \cite{itut2025}.

After generating the action plan, the agent submits a natural language query to the AgentDNS root server to discover suitable third-party services. For instance, in Step 1, the agent sends the \texttt{tool\_function} description directly to AgentDNS, which uses intelligent retrieval methods to identify matching services. Suppose AgentDNS returns a service named \texttt{agentdns://example/search/searchagent}; it also provides metadata such as the physical endpoint, supported protocols, service cost, capabilities, and available APIs. The agent uses this information to invoke the selected third-party service.

Following service discovery, the agent enters the action execution phase. During this stage, it executes the steps of the action plan in sequence. When a step requires a service, the agent uses the corresponding protocol to access the third-party service obtained from AgentDNS and passes the result to the next step. For steps that do not involve external services, the agent inputs the step purpose description, along with previous outputs and prompt instructions, into the LLM for generation. This process continues until all steps in the action plan are completed.

This case study presents a simplified example, while in practice, the structure and format of an action plan can be adapted to suit different needs. Importantly, the third-party service descriptions within the action plan are expressed in natural language, which means they are not tightly coupled with specific service identifiers, tool names, or endpoint URLs. AgentDNS plays a critical role in decoupling the foundational agent model from vendor-specific details such as service names, tool identifiers, and physical addresses, enabling more flexible and scalable agent architectures.

\section{Future opportunities}
While AgentDNS addresses fundamental challenges in service discovery, interoperability, and billing in the agent ecosystem, numerous directions remain open for future exploration. These include decentralized and federated architecture, AgentDNS-compatible agent planning LLMs, privacy-preserving and trusted discovery, as well as AgentDNS service discovery optimization. First, while the current design of AgentDNS adopts a centralized architecture, future iterations may benefit from decentralized or federated \cite{huang2024aggregate} architecture, such as blockchain \cite{li2021b,karaarslan2018blockchain}. This would improve robustness, reduce the risk of single points of failure, and enhance trust in cross-organizational collaborations. Second, training and fine-tuning agent planning LLMs \cite{wang2023describe, hu2024agentgen} specifically compatible with AgentDNS is also an important direction. This can involve constructing agent planning datasets and fine-tuning LLMs to enhance their compatibility with AgentDNS. Alternatively, reinforcement learning techniques \cite{wen2024reinforcing, jin2025search, qi2024webrl, peiyuan2024agile} can be used to train agents to autonomously explore and optimize action sequences, dynamically selecting and combining various services registered in AgentDNS to maximize task success rates and efficiency. Third, security and privacy will remain central in cross-vendor agent collaboration. Future directions may involve privacy-preserving search and resolution, using technologies such as homomorphic encryption\cite{buban2025encrypted}, differential privacy, and secure multi-party computation. AgentDNS could also integrate trust and reputation systems to allow agents to evaluate service quality and security risks before invocation. Finally, the optimization of AgentDNS service discovery and retrieval remains a critical area for improving system performance and user experience. 

\section{Conclusion}
The rapid advancement of LLM agents has exposed critical gaps in cross-vendor service discovery, interoperability, and authentication, hindering the vision of autonomous multi-agent collaboration. This paper introduces AgentDNS, a unified root domain naming system designed to bridge these gaps by providing a semantically rich namespace, natural language-driven service discovery, protocol-aware interoperability, and trustless authentication and billing. By decoupling agent identifiers from physical addresses and embedding dynamic metadata resolution, AgentDNS enables agents to autonomously discover, resolve, and securely invoke services across organizational and technological boundaries. Our architecture and case studies demonstrate its potential to streamline multi-agent workflows, reduce manual overhead, and foster an open ecosystem for agent collaboration. Future works include decentralized and federated architecture, AgentDNS-compatible agent planning LLMs, privacy-preserving and trusted discovery, as well as AgentDNS service discovery optimization, etc.

\section{Acknowledgments}
This work was supported by the National Key R\&D Program of China under Grant No.2023YFB2904100.

\appendix

\bibliography{aaai25.bib}

\end{document}